\documentclass[11pt]{article}
\usepackage{microtype}
\usepackage{graphicx}
\usepackage{subcaption}
\usepackage{pgf}
\usepackage{enumitem}
\usepackage{booktabs} 
\usepackage{comment}
\usepackage{hyperref}
\usepackage{amsmath}
\usepackage{amssymb}
\usepackage{amsthm}
\usepackage{bm}
\usepackage{mathtools}
\usepackage{commath}
\usepackage{times}
\usepackage{latexsym}
\usepackage{xcolor}
\usepackage{adjustbox}
\usepackage{graphicx}
\usepackage{booktabs}
\usepackage{graphics}
\usepackage{pifont}
\newcommand{\cmark}{\ding{51}}%
\newcommand{\xmark}{\ding{55}}%
\usepackage{multirow,multicol}

\usepackage{scalerel}[2014/03/10]
\usepackage{stackengine,wasysym}

\usepackage{geometry}
\newcommand\wtil[1]{\ThisStyle{%
  \setbox0=\hbox{$\SavedStyle#1$}%
  \stackengine{-.1\LMpt}{$\SavedStyle#1$}{%
    \stretchto{\scaleto{\SavedStyle\mkern.2mu\sim}{.5467\wd0}}{.7\ht0}
  }{O}{c}{F}{T}{S}%
}}

\def\1{\bm{1}}

\def\vtheta{{\bm{\theta}}}
\def\va{{\bm{a}}}
\def\vb{{\bm{b}}}

\def\vh{{\bm{h}}}

\def\vk{{\bm{k}}}

\def\vp{{\bm{p}}}
\def\vq{{\bm{q}}}

\def\vv{{\bm{v}}}
\def\vw{{\bm{w}}}
\def\vx{{\bm{x}}}

\def\vz{{\bm{z}}}

\def\mK{{\bm{K}}}

\def\mQ{{\bm{Q}}}

\def\mV{{\bm{V}}}
\def\mW{{\bm{W}}}

\DeclareMathAlphabet{\mathsfit}{\encodingdefault}{\sfdefault}{m}{sl}
\SetMathAlphabet{\mathsfit}{bold}{\encodingdefault}{\sfdefault}{bx}{n}

\def\sR{{\mathbb{R}}}

\newcommand{\R}{\mathbb{R}}

\title{Learning to Encode Position for Transformer \\ with Continuous Dynamical Model}
\author{
\small Xuanqing Liu$^\dag$,\ Hsiang-Fu Yu$^\ddag$,\ Inderjit Dhillon$^{\S\ddag}$,\ Cho-Jui Hsieh$^\dag$\\
\small $^\dag$ UCLA\quad $\S$ UT Austin\quad $\ddag$ Amazon Inc.\\
\small \href{mailto:xqliu@cs.ucla.edu}{xqliu@cs.ucla.edu}\quad \href{mailto:rofu.yu@gmail.com}{rofu.yu@gmail.com}\\ 
\small \href{mailto:inderjit@cs.utexas.edu}{inderjit@cs.utexas.edu}\quad \href{mailto:chohsieh@cs.ucla.edu}{chohsieh@cs.ucla.edu} \\
}

\date{}
\begin{document}
\maketitle
\begin{abstract}
 We introduce a new way of learning to encode position information for non-recurrent models, such as Transformer models. Unlike RNN and LSTM, which contain inductive bias by loading the input tokens sequentially, non-recurrent models are less sensitive to position. The main reason is that position information among input units is not inherently encoded, i.e., the models are permutation equivalent; this problem justifies why all of the existing models are accompanied by a sinusoidal encoding/embedding layer at the input. However, this solution has clear limitations: the sinusoidal encoding is not flexible enough as it is manually designed and does not contain any learnable parameters, whereas the position embedding restricts the maximum length of input sequences. It is thus desirable to design a new position layer that contains learnable parameters to adjust to different datasets and different architectures. At the same time, we would also like the encodings to extrapolate in accordance with the variable length of inputs. In our proposed solution, we borrow from the recent Neural ODE approach, which may be viewed as a versatile continuous version of a ResNet. This model is capable of modeling many kinds of dynamical systems. We model the evolution of encoded results along position index by such a dynamical system, thereby overcoming the above limitations of existing methods. We evaluate our new position layers on a variety of neural machine translation and language understanding tasks, the experimental results show consistent improvements over the baselines.
\end{abstract}
\section{Introduction}
\label{sec:intro}
Transformer based models~\cite{vaswani2017attention,devlin2018bert,yang2019xlnet,radford2019language, lan2019albert,raffel2019exploring}
have become one of the most effective approaches to model sequence data of
variable lengths. Transformers have shown wide applicability to many natural
language processing (NLP) tasks such as language
modeling~\cite{radford2019language}, neural machine translation
(NMT)~\cite{vaswani2017attention}, and language
understanding~\cite{devlin2018bert}.  Unlike traditional recurrent-based
models (e.g., RNN or LSTM), Transformer utilizes a non-recurrent but
self-attentive neural architecture to model the dependency among elements at
different positions in the sequence, which leads to better parallelization
using modern hardware and alleviates the vanishing/exploding gradient problem
in traditional recurrent models.
\par
\cite{yun2019Transformers} prove that the design of self-attentive architecture leads to a family of permutation equivalence functions. Thus, for applications where the ordering of the elements matters, how to properly encode position information is crucial for  Transformer based models. There have been many attempts to encode position information for the Transformer. In the original Transformer paper~\cite{vaswani2017attention}, a family of pre-defined sinusoidal functions was adapted to construct a set of embeddings for each position. These fixed position embeddings are then added to the word embeddings of the input sequence accordingly. To further construct these position embeddings in a more data-driven way, many recent Transformer variants such as \cite{devlin2018bert, liu2019roberta} include these embeddings as learnable model parameters in the training stage. This data-driven approach comes at the cost of the limitation of a fixed maximum length of input sequence $L_{\text{max}}$ and the computational/memory overhead of additional $L_{\text{max}}\times d$ parameters, where $L_{\text{max}}$ is usually set to 512 in many applications, and $d$ is the dimension of the embeddings. \cite{shaw2018self} propose a relative position representation to reduce the number of parameters to $(2K+1) d$ by dropping the interactions between tokens with a distance greater than $K$. In addition to just the input layer, \cite{dehghani2018universal} and \cite{lan2019albert} suggest that the injection of position information to every layer leads to even better performance for the Transformer.

An ideal position encoding approach should satisfy the following three properties:
\begin{enumerate}[nosep,leftmargin=1em,labelwidth=*,align=left]
    \item {\bf Inductive}: the ability to handle sequences longer than any sequence seen in the training time.
    \item {\bf Data-Driven}: the position encoding should be learnable from the data.
    \item {\bf Parameter Efficient}: number of trainable parameters introduced by the encoding should be limited to avoid increased model size, which could hurt generalization.
\end{enumerate}

In Table~\ref{tab:capabilities}, we summarize some of the existing position encoding approaches in terms of these three properties.
\par

In this paper, we propose a new method to encode position with minimum cost. The main idea is to model position encoding as a continuous dynamical system, so we only need to learn the system dynamics instead of learning the embeddings for each position independently. By doing so, our method enjoys the best of both worlds -- we bring back the inductive bias, and the encoding method is freely trainable while being parameter efficient. To enable training of this dynamical system with backpropagation, we adopt the recent progress in continuous neural network~\cite{chen2018neural}, officially called Neural ODE. In some generative modeling literature, it is also called the free-form flow model~\cite{grathwohl2018ffjord}, so we call our model \textbf{FLOw-bAsed TransformER (FLOATER)}. We highlight our contributions as follows:
\begin{itemize}[nosep,leftmargin=1em,labelwidth=*,align=left]
    \item We propose FLOATER, a new position encoder for Transformer, which models the position information via a continuous dynamical model in a data-driven and parameter-efficient manner.
    \item Due to the use of a continuous dynamic model, FLOATER can handle sequences of any length. This property makes inference more flexible.
    \item With careful design, our position encoder is {\bf compatible} with the original Transformer; \emph{i.e.}, the original Transformer can be regarded as a special case of our proposed position encoding approach. As a result, we are not only able to train a Transformer model with FLOATER from scratch but also plug FLOATER into most existing pre-trained Transformer models such as BERT, RoBERTa, \emph{etc.}
    \item We demonstrate that FLOATER consistent improvements over baseline models across a variety of NLP tasks ranging from machine translations, language understanding, and question answering.
\end{itemize}
\begin{table}[tb!]
    \centering
    \caption{Comparing position representation methods}
    \label{tab:capabilities}
    \begin{adjustbox}{width=0.7\columnwidth,center}
    \begin{tabular}{lccc}
    \toprule
    Methods       & Inductive & Data-Driven & Parameter Efficient \\
    \midrule
    Sinusoidal~\cite{vaswani2017attention}    & \cmark & \xmark & \cmark \\
    Embedding~\cite{devlin2018bert}     & \xmark & \cmark & \xmark \\
    Relative~\cite{shaw2018self}      & \xmark & \cmark & \cmark \\
    This paper    & \cmark & \cmark & \cmark \\
    \bottomrule
    \end{tabular}
    \end{adjustbox}
\end{table}
\section{Background and Related Work}
\label{sec:background}
\subsection{Importance of Position Encoding for Transformer}
\label{sec:importance}
We use a simplified self-attentive sequence encoder to illustrate the
importance of position encoding in the Transformer. Without position
encoding, the Transformer architecture can be viewed as a stack of $N$
blocks $B_n: n= 1,\ldots, N$ containing a self-attentive $A_n$ and a
feed-forward layer $F_n$. By dropping the residual connections and layer
normalization, the architecture of a simplified Transformer encoder can be
represented as follows.
\begin{align}
    \label{eq:simple-Transformer-encoder}
    \text{Encode}(\vx) &=B_{N}\circ B_{N-1}\circ\cdots \circ B_1(\vx), \\ 
    B_{n}(\vx)&= F_{n}\circ A_n\left(\vx\right) \label{eq:building-block},
\end{align}
where $\vx =[\vx_1,\vx_2,\dots,\vx_L]^\top \in \R^{L\times d}$, $L$ is the
length of the sequence and $d$ is the dimension of the word embedding.
$A_n(\cdot)$ and $F_n(\cdot)$ are the self-attentive and feed-forward layer in the $n$-th block $B_n(\cdot)$, respectively.

Each row of $A_1(\vx)$ can be regarded as a weighted sum of the value matrix $\mV\in\sR^{L\times d}$, with the weights determined by similarity
scores between the key matrix $\mK\in\sR^{L\times d}$ and query matrix $\mQ\in\sR^{L\times d}$ as follows:
\begin{equation}
    \label{eq:self-attention}
    \begin{aligned}
    &A_1(\vx)=\texttt{Softmax}\Big(\frac{\mQ\mK^\top}{\sqrt{d}}\Big)\mV,\\
    &\begin{aligned}
   \mQ &= [\vq_1,\vq_2,...,\vq_L]^\top,\quad &\vq_i= \mW_q \vx_i + \vb_q,\\
   \mK &= [\vk_1,\vk_2,...,\vk_L]^\top,\quad &\vk_i= \mW_k \vx_i + \vb_k, \\
   \mV &= [\vv_1, \vv_2,...,\vv_L]^\top,\quad &\vv_i= \mW_v \vx_i + \vb_v, \\
   \end{aligned} 
    \end{aligned}
\end{equation}
$\mW_{q/k/v}$ and $\vb_{q/k/v}$ are the weight and bias parameters introduced in the self-attentive function $A_1(\cdot)$. 
The output of the feed-forward function $F_1(\cdot)$ used in the Transformer is also a matrix with $L$ rows. In particular, the $i$-th row is obtained as follows.
\begin{equation}
    \label{eq:ffn}
    \text{the $i$-th row of } F_1(\vx) = \mW_2 \sigma(\mW_1\vx_i + \vb_1) +\vb_2,
\end{equation}
where $\mW_{1,2}$ and $\vb_{1,2}$ are the weights and biases of linear transforms, and $\sigma(\cdot)$ is the activation function. It is not hard to see from \eqref{eq:self-attention} and \eqref{eq:ffn} that both $A_1(\cdot)$ and $F_1(\cdot)$ are permutation equivalent. Thus, we can conclude that the entire function defined in \eqref{eq:simple-Transformer-encoder} is also permutation equivalent, i.e., $\Pi\times \text{Encode}(\vx) = \text{Encode} \left(\Pi \times \vx\right)$ for any $L\times L$ permutation matrix $\Pi$. 
This permutation equivalence property restricts the Transformer without position information from modeling sequences where the ordering of elements matters. 

\begin{figure*}[htb]
    \centering
    \includegraphics[width=0.8\linewidth]{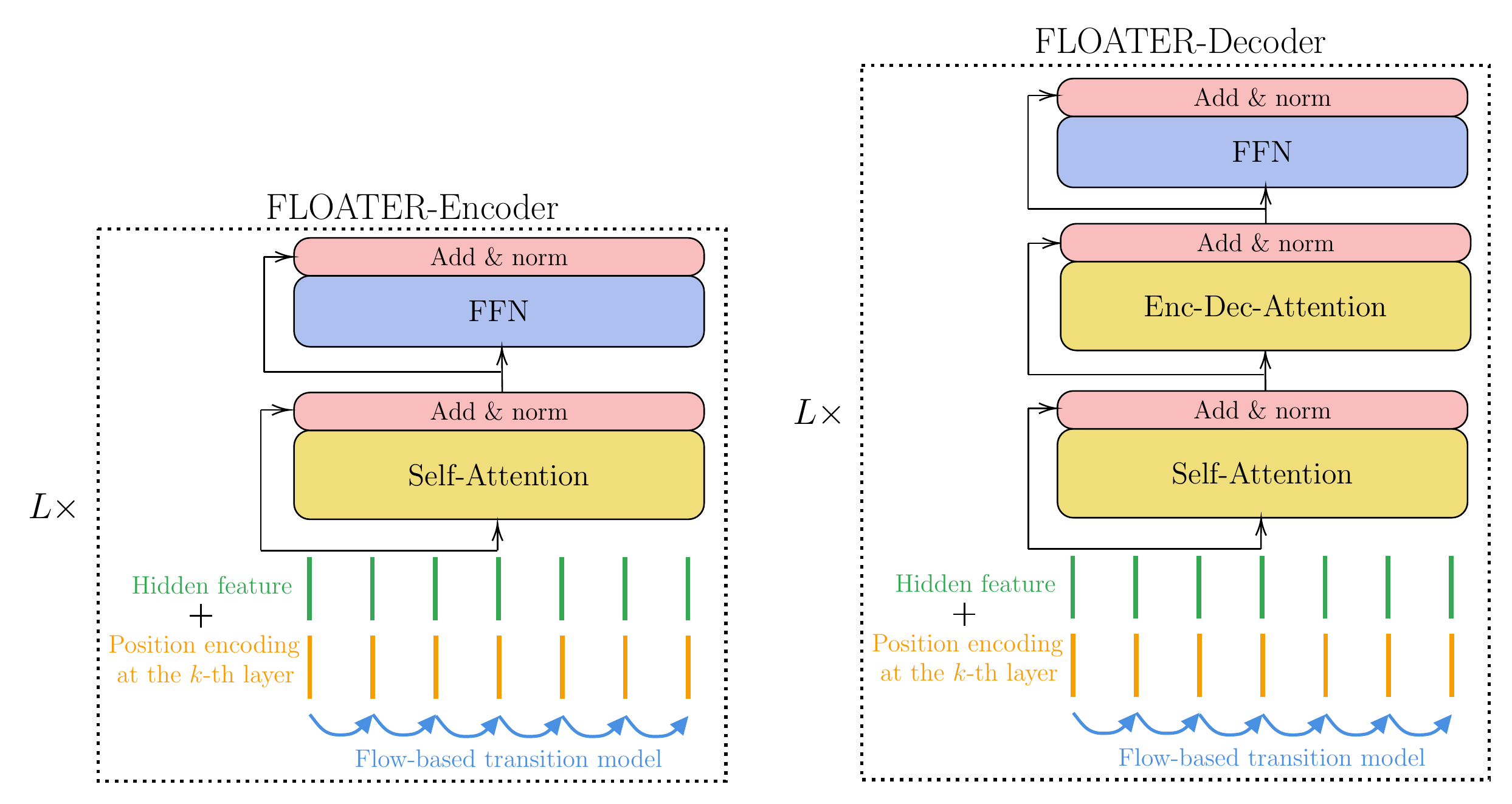}
    \caption{The architecture of our model (FLOATER). The main differences between FLOATER and the original Transformer model are: 1) the position representation is integrated into each block in the hierarchy (there are $N$ blocks in total); and 2) there is a dynamical model (see \eqref{eq:dynamics}) that generates position encoding vectors for each block. The dynamics are solved with a black-box ODE solver detailed in the supplementary material. }
    \label{fig:arch-flow-Transformer}
\end{figure*}
\subsection{Position Encoding in Transformer}
\label{sec:pos-enc-formulation}
As mentioned in Section~\ref{sec:intro}, there are many attempts to inject position information in self-attentive components. Most of them can be described in the following form:
\begin{align}
    B_{n}(\vx) = F_n \circ A_n \circ \Phi_n (\vx),\ n \in \{1,...,N\}, 
\end{align}
where $\Phi_n(\vx)$ is a position encoding function.  

\cite{vaswani2017attention} propose to keep $\Phi_{n}(\vx)=\vx, \forall n \ge 2$ and inject position information only at the input block with a family of pre-defined sinusoidal functions: $\Phi_1(\vx) = \vx + \vp^{(1)}$, where $\vp^{(1)} = [\vp^{(1)}_1, \vp^{(1)}_2, ..., \vp^{(1)}_L$] is a position embedding matrix with the $i$-th row corresponding to the $i$-th position in the input sequence. In particular, the $j$-th dimension of the $i$-th row is defined as follows. 
\begin{align}\label{eq:sin-encoding}
    \vp^{(1)}_i[j] &= 
    \begin{cases}
    \sin(i\cdot c^{\frac{j}{d}}) & \text{if $j$ is even}, \\
    \cos(i\cdot c^{\frac{j-1}{d}}) & \text{if $j$ is odd},
    \end{cases}
\end{align} 
where $c=10^{-4}$.
\cite{dehghani2018universal} and \cite{lan2019albert} observe better performance by further injecting the position information at each block, i.e., $\Phi_n(\vx) = \vx + \vp^{(n)}$ as follows:
\begin{align}\label{eq:universal-position-encoding}
    \vp^{(n)}_i[j]\vspace{-1em} &= \vspace{-1em}
    \begin{cases} 
    \sin(i\cdot {c}^{\frac{j}{d}}) + 
    \sin(n\cdot {c}^{\frac{j}{d}}) & \text{if $j$ is even}, \\
    \cos(i\cdot {c}^{\frac{j-1}{d}}) + 
    \cos(n\cdot {c}^{\frac{j-1}{d}}) & \text{if $j$ is odd}.
    \end{cases}
\end{align}
Note that for the above two approaches, position encoding functions $\Phi_n(\cdot)$ are fixed for all the applications. Although no additional parameters are introduced in the model, both approaches are inductive and can handle input sequences of variable length. 

Many successful variants of pre-trained Transformer models, such as BERT~\cite{devlin2018bert} and RoBERTa~\cite{liu2019roberta}, include the entire embedding matrix $\vp^{(1)}\in \sR^{L\times d}$ in $\Phi_1(\vx)$ as training parameters. As the number of training parameters needs to be fixed, the maximum length of a sequence, $L_{\text{max}}$, is required to be determined before the training. Although it lacks the inductive property, this data-driven approach is found to be effective for many NLP tasks. Note that, unlike the fixed sinusoidal position encoding, there is no attempt to inject a learnable position embedding matrix at each block for Transformer due to a large number of additional parameters ($NL_{\text{max}}d$). 

\section{FLOATER: Our Proposed Position Encoder}
\label{sec:flower}
We introduce our method in three steps. In the first step, we only look at one Transformer block, and describe how to learn the position representation driven by a dynamical system; in the second step, we show how to save parameters if we add position signals to every layer; lastly, we slightly change the architecture to save trainable parameters further and make FLOATER ``compatible'' with the original Transformer~\cite{vaswani2017attention}. The compatibility means our model is a strict superset of the vanilla Transformer so that it can be initialized from the Transformer. 
\subsection{Position Encoding with Dynamical Systems}
\label{sec:modeling}
Position representations in Transformer models are a sequence of 
vectors $\{\vp_i \in \sR^d: i = 1,...,L\}$ to be added to the sequence 
of the input representations \{$\vx_i:i=1,...,L\}$. 
Existing position encoding approaches either apply a fixed sinusoidal function to obtain $\{\vp_i\}$, or include them as uncorrelated learnable parameters. 
Both of them fail to capture the dependency or dynamics among these position representations $\{\vp_i\}$. 
In this paper, we propose to use a dynamical system to model these position representations; that is, 
there is a ``latent force'' denoted by $\vh_i$ that drives the changes from $\vp_i$ to $\vp_{i+1}$.  
To encourage smoothness, we consider $\vp(t):\sR_+\mapsto \sR^d$ as the continuous version of the 
discrete sequence $\{\vp_i\}$. In particular, our proposed continuous dynamical system is characterized as follows:
\begin{equation}
    \vp(t)\!=\!\vp(s)\!+\!\int_s^t\vh(\tau, \vp({\tau});\vtheta_h)\dif \tau,\ 0\le s\le t<\infty,
    \label{eq:dynamics}
\end{equation}
together with an initial vector $\vp(0)$, where $\vh(\tau, \vp({\tau});\vtheta_h)$ is a neural network 
parameterized by $\vtheta_h$ and takes the previous state $(\tau, \vp(\tau))$. Notice that the domain of 
$\vp(\cdot)$ is $\sR_+$. The position sequence $\{\vp_i\}$ can be obtained by taking $\vp(\cdot)$ on a series of points $\{t_i: 0 \le t_1 < \cdots \le t_L\}$: $\vp_i=\vp(t_i)$.
One simple strategy is to set $t_i=i\cdot\Delta t$ so that the points are equidistant, where $\Delta$ is a hyperparameter (e.g., $\Delta=0.1$). 
With this strategy, we are implicitly assuming the position signals evolve steadily as we go through each token in a sentence. 
In general, $\{t_i\}$ can be any monotonically increasing series, which allows us to extend our work to more applications where the elements in the sequence are not always observed with the same interval. More discussions about the applicability for this general setting is included in the Supplementary material. For the NLP applications discussed in this paper, we choose $t_i=i\cdot \Delta t$.
\par
Eq.~\eqref{eq:dynamics} is equivalent to an ODE problem $\frac{\dif \vp(t)}{\dif t}=\vh(t,\vp(t);\vtheta_h)$, which is guaranteed to have a unique solution under mild conditions~\cite{clark1964ordinary}. We follow the efficient approach by \cite{chen2018neural} to calculate the gradients of $\vtheta_h$ with respect to the overall training loss, which allows us to include this parameterized dynamical position encoder into the end-to-end training of Transformer models. More details can be found in the Supplementary material.
\par
Our dynamical system~\eqref{eq:dynamics} is quite flexible to admit the standard sinusoidal position encoding~\eqref{eq:sin-encoding} as a special case: 
\begin{equation}
    \label{eq:transformer-pos-enc-flow}
    \begin{aligned}
    & \vp_{i+1}[j] - \vp_{i}[j] \\
    &=\begin{cases}
    \sin\big((i+1)\cdot c^{\frac{j}{d}}\big)-\sin\big(i\cdot c^{\frac{j}{d}}\big) & \text{if $j$ is even}\\
    \cos\big((i+1)\cdot c^{\frac{j-1}{d}}\big)-\cos\big(i\cdot c^{\frac{j-1}{d}}\big)& \text{if $j$ is odd}\\
    \end{cases}\\
    &=
    \begin{cases}
    \int_i^{i+1}c^{-\frac{j}{d}}\cos(\tau\cdot  c^{\frac{j}{d}})\dif \tau & \text{ if $j$ is even}\\
    \int_i^{i+1}-c^{-\frac{j-1}{d}}\sin(\tau\cdot  c^{\frac{j-1}{d}})\dif \tau& \text{ if $j$ is odd}, 
    \end{cases}
    \end{aligned}
\end{equation}
This indicates that for simple sinusoidal encoding, there exists a dynamical system $\vh(\cdot)$ which is also sinusoidal function.
\par

\subsection{Parameter Sharing among Blocks}
As mentioned in Section~\ref{sec:background}, injecting position information to each block for Transformer leads to better performance~\cite{dehghani2018universal,lan2019albert} in some language understanding tasks. Our proposed position encoder FLOATER~\eqref{eq:dynamics} can also be injected into each block. The idea is illustrated in Figure~\ref{fig:arch-flow-Transformer}. 
Typically there are $6$ blocks in sequence-to-sequence Transformer and $12$ or $24$ blocks in BERT. We add a superscript $(n)$ to denote dynamics at $n$-th block: 
\begin{equation*}
    \vp^{(n)}(t) = \vp^{(n)}(s) + \int_s^t\vh^{(n)}(\tau, \vp^{(n)}(\tau);\vtheta_h^{(n)})\dif \tau.
\end{equation*}
As we can imagine, having $N$ different dynamical models $\vh^{(n)}(\cdot;\vtheta_h^{(n)})$ for each block can introduce too many parameters and  cause significant training overhead. Instead, we address this issue by sharing parameters across all the blocks, namely
\begin{equation}
    \label{eq:sharing-w}
    \vtheta_h^{(1)}=\vtheta_h^{(2)}=\dots=\vtheta_h^{(N)}.
\end{equation}
Note that \eqref{eq:sharing-w} does not imply that all the $\vp_t^{(n)}$ are the same, as we will assign different initial values for each block, that is  $\vp^{(n_1)}(0)\ne\vp^{(n_2)}(0)$ for $n_1 \ne n_2$.
\begin{table*}[htb]
    \centering
    \begin{tabular}{lcccc}
    \toprule
       &\multicolumn{2}{c}{\textit{Transformer-Base}} & \multicolumn{2}{c}{\textit{Transformer-Large}}  \\  
       & En-De & En-Fr & En-De & En-Fr  \\
       \midrule
        \multicolumn{5}{l}{\bf Position encoders at all blocks} \\
        FLOATER & \textbf{28.6} & \textbf{41.6} & \textbf{29.2} & \textbf{42.7}\\
        Pre-defined Sinusoidal Position Encoder & 28.2 & 40.6 & 28.4 & 42.0 \\
        Fixed-length Position Embedding & 26.9 & 40.9 & 28.3& 42.0 \\
       \midrule
        \multicolumn{5}{l}{\bf Position encoder only at input block} \\
        FLOATER & 28.3 & 41.1 & 29.1 & 42.4 \\
        Pre-defined Sinusoidal Position Encoder & 27.9 & 40.4 & 28.4 & 41.8 \\
        Fixed-length Position Embedding & 27.8 & 40.9 & 28.5 & 42.4 \\
        \bottomrule
    \end{tabular}
    \caption{Experimental results of various position encoders on the machine translation task.} 
    \label{tab:exp-NMT}
\end{table*}
\subsection{Compatibility and Warm-start Training\label{sec:compatible}}
In this section, we change the way to add position encoding so that our FLOATER can be directly initialized from Transformer. As an example, we use the standard Transformer model, which has a fixed sinusoidal encoding at the input block and no position encoding at deeper levels. Note that this technique can be extended to other variants of Transformers with different position encoding methods, such as embedding matrix. We first examine the standard Transformer model, the query matrix $\mQ^{(n)}$ at block-$n$ is
\begin{align}
\wtil{\vq}_i^{(n)}&=\mW_{q}^{(n)}\big(\vx_i+\wtil{\vp}^{(n)}_i\big)+\vb_{q}^{(n)},\label{subeq:transformer-n1}
\end{align}
where $\mW^{(n)}_q$ and $\vb^{(n)}_q$ are parameters in $A_n$~\eqref{eq:self-attention}; $\wtil{\vp}^{(n)}$ is the sinusoidal encoding; $\wtil{\vq}_i^{(n)}$ is the $i$-th row of $\mQ^{(n)}$. Here we add a tilde sign to indicate the sinusoidal vectors. Formulas for $\wtil{\vk}_i^{(n)}$ and $\wtil{\vv}_i^{(n)}$ have a very similar form and are omitted for brevity.

Now we consider the case of FLOATER, where new position encodings $\vp_i$ are added
\begin{equation}
\label{eq:add-pos-bias-input}
\begin{aligned}
\vq_i^{(n)} &=\mW^{(n)}_{q}\big(\vx_i+\vp_i\big)+\vb^{(n)}_{q}\\
&=\underbrace{\mW^{(n)}_q(\vx_i+\wtil{\vp}^{(n)}_i)+\vb^{(n)}_q}_{\text{Eq. }\eqref{subeq:transformer-n1}}+\underbrace{\mW^{(n)}_{q}(\vp_{\textcolor{red}{i}}-\wtil{\vp}^{(n)}_{\textcolor{red}{i}})}_{\text{Extra bias term depends on $\color{red}i$}}\\
&=\wtil{\vq}^{(n)}_i+{\vb}^{(n)}_{q,\textcolor{red}{i}}.
\end{aligned}
\end{equation}
It is easy to see that the changing the position embedding from $\{\wtil{\vp}_i^{(n)}\}$ to $\{\vp_i^{(n)}\}$ is equivalent to adding a position-aware bias vector $\vb^{(n)}_{q, \textcolor{red}{i}}$ into each self-attentive layers $\{A_n(\cdot)\}$. As a result, we can instead apply \eqref{eq:dynamics} to model the dynamics of $\vb_q^{(n)}$. In particular, we have the following dynamical system:
\begin{equation}\label{eq:bias-dynamics}
    \vb_{q}^{(n)}(t)=\vb^{(n)}_{q}(0) + \int_0^t\vh^{(n)}(\tau, \vb^{(n)}_{q}({\tau}); \vtheta_h)\dif \tau.
\end{equation}
After that, we set $\vb^{(n)}_{q,\textcolor{red}{i}} = \vb^{(n)}_q(\textcolor{red}{i} \cdot \Delta t)$. 
We can see that if  $\vh(\cdot)=0$ and $\vb^{(n)}_{q}(0)=0$, then $\vb_q^{(n)}\equiv 0$. This implies \eqref{eq:add-pos-bias-input} degenerates to \eqref{subeq:transformer-n1}. 
Note that \eqref{eq:bias-dynamics} has the same form as \eqref{eq:dynamics}, except that we are now modeling the bias terms $\vb_{q,i}$ in~\eqref{eq:self-attention}. We will apply the same technique to $\mK$ and $\mV$.
\par
To summarize, our model has a tight connection to the original Transformer: if we set all dynamical models to zero, which means $\vh(\tau,\vp(\tau);\vtheta_h)\equiv 0$, then our FLOATER model will be equivalent to the original Transformer with the sinusoidal encoding. The same trick also works for Transformer with position embedding such as BERT~\cite{devlin2018bert}.
\par
We strive to make our model compatible with the original Transformer due to the following reasons. First of all, the original Transformer is faster to train as it does not contain any recurrent computation; this is in contrast to our dynamical model~\eqref{eq:dynamics}, where the next position $\vp_{i+1}$ depends on the previous one $\vp_{i}$. By leveraging the compatibility of model architecture, we can directly initialize FLOATER model from a pre-trained Transformer model checkpoint and then fine-tune for the downstream task for a few more epochs. By doing so, we enjoy all the benefits of our FLOATER model but still maintain an acceptable training budget. Likewise, for models such as BERT or Transformer-XL, we already have well-organized checkpoints out of the box for downstream tasks. These models are costly to train from scratch, and since our goal is to examine whether our proposed position representation method can improve over the original one, we decided to copy the weights layer by layer for attention as well as FFN layers, and randomly initialize the dynamical model $\vh(\tau, \vp(\tau); \vtheta_h)$.

\section{Experimental Results}
\label{sec:exp}
In this section, we perform experiments to see if FLOATER can improve over the existing position encoding approaches for a given Transformer model on various NLP tasks. 
Thus, all the metrics reported in this paper are computed from a single (not ensemble) Transformer model over each evaluation NLP task. Albeit lower than top scores on the leaderboard, these metrics are able to reveal more clear signal to judge the effectiveness of the proposed position encoder.   

All our codes to perform experiments in this paper are based on the Transformer implementations in the \texttt{fairseq}~\cite{ott2019fairseq} package. Implementation details can be found in the Supplementary material. Our experimental codes will be made publicly available. 

\begin{table*}[htb]
    \centering
    \caption{Experimental results on GLUE benchmark}
    \label{tab:exp-GLUE}
    \resizebox{0.9\textwidth}{!}{%
    \begin{tabular}{@{}lcccccccc@{}}
    \toprule
    \multirow{2}{*}{Model} & \multicolumn{2}{c}{\textbf{Single Sentence}} & \multicolumn{3}{c}{\textbf{Similarity and Paraphrase}} & \multicolumn{3}{c}{\textbf{Natural Language Inference}} \\
    & CoLA & SST-2 & MRPC & QQP & STS-B &  MNLI & QNLI & RTE\\
    \midrule
    \multicolumn{9}{l}{\textit{Base model}}\\
    RoBERTa & \textbf{63.6} & 94.8 & 88.2 & \textbf{91.9} & 91.2 & 87.6 & 92.8 & 78.7 \\
    FLOATER & 63.4 & \textbf{95.1} & \textbf{89.0} & 91.7 & \textbf{91.5} & \textbf{87.7} & \textbf{93.1} & \textbf{80.5} \\
    \midrule
    \multicolumn{9}{l}{\textit{Large model}}\\
    RoBERTa & 68.0 & 96.4 & 90.9 & \textbf{92.2} & 92.4 & 90.2 & 94.7 & 86.6 \\
    FLOATER & \textbf{69.0} & \textbf{96.7} & \textbf{91.4} & \textbf{92.2} & \textbf{92.5} & \textbf{90.4} & \textbf{94.8} & \textbf{87.0} \\
    \bottomrule
    \end{tabular}
    } 
\end{table*}

\begin{table}[htb]
    \centering
    \caption{Experiment results on RACE benchmark. ``Middle'' means middle school level English exams, ``High'' means high school exams. Other details can be found in \cite{lai2017race}.}
    \label{tab:race-benchmark}
    \begin{tabular}{lccc}
    \toprule
    Model & Accuracy & Middle & High \\
    \midrule
    \multicolumn{4}{l}{\textit{Single model on test, large model}}\\
    RoBERTa      & 82.8  & 86.5 & 81.3 \\
    FLOATER & \textbf{83.3}  & \textbf{87.1} & \textbf{81.7} \\
    \bottomrule
    \end{tabular}
\end{table}
\subsection{Neural Machine Translation}
\label{sec:exp-nmt}
Neural Machine Translation (NMT) is the first application that demonstrates the superiority of a sequence-to-sequence Transformer model over conventional recurrent sequence models. 
We include the following three additive position encoders: $\Phi^{(n)}(\vx) = \vx + \vp^{(n)}$.
\begin{itemize}[nosep,leftmargin=1em,labelwidth=*,align=left]
\item \textbf{Data-driven FLOATER}: $\vp^{(n)}$ is generated by our proposed continuous dynamical models with data-driven parameters described in \eqref{eq:dynamics}. 
\item \textbf{Pre-defined sinusoidal position encoder}: $\vp^{(n)}$ is constructed by a pre-defined function described in $\eqref{eq:universal-position-encoding}$, which is proposed by \cite{vaswani2017attention} and extended by \cite{dehghani2018universal}.  
\item \textbf{Length-fixed position embedding}: $\vp^{(n)}$ is included as learnable training parameters. This is first introduced by \cite{vaswani2017attention} and adopted in many variants of Transformer~\cite{devlin2018bert,liu2019roberta}. 
\end{itemize}
To better demonstrate the parameter efficiency brought by FLOATER, for each above encoder, we also include two experimental settings: position encoder at all blocks or only at the input block (i.e., $\vp^{(n)} = 0, \forall n \ge 2$).

In Table~\ref{tab:exp-NMT}, we present the BLEU scores on WMT14 Ee-De and En-Fr datasets with both \textit{Transformer-base} and \textit{Transformer-large} models described in \cite{vaswani2017attention}.
Among all the data/model combinations, our proposed FLOATER at all blocks outperforms two other position encoders. 

On the other hand, we also observe that adding position encoders at all blocks yields better performance than only at the input block. While there is an exception in the fixed-length position embedding approach. We suspect that this phenomenon is due to over-fitting cased by $L_{\text{max}}dN$ learnable parameters introduced by this approach. In contrast, our proposed FLOATER is parameter efficient (more discussions in Section~\ref{sec:exp-analysis}), so the performance can be improved by injecting the position encoder at all the blocks of Transformer without much additional overhead. 

\subsection{Language Understanding and Question Answering}
\label{sec:exp-luqa}
\begin{table}[htb]
    \centering
    \caption{Experiment results on SQuAD benchmark. All results are obtained from RoBERTa-large model.\label{tab:squad}}
    \begin{tabular}{lcccc}
    \toprule
    \multirow{2}{*}{Model}     &  \multicolumn{2}{c}{\textbf{SQuAD 1.1}} & \multicolumn{2}{c}{\textbf{SQuAD 2.0}}\\
    & EM & F1 & EM & F1 \\
    \midrule
    \multicolumn{5}{l}{\textit{Single models on dev, w/o data augmentation}}\\
    RoBERTa   & 88.9 & 94.6 & 86.5 & 89.4 \\
    FLOATER    & 88.9 & 94.6 & \textbf{86.6} & \textbf{89.5} \\
    \bottomrule
    \end{tabular}
\end{table}
Pretrained Transformer models such as BERT and RoBERTa have become the key to achieving the state-of-the-art performance for various language understanding and question answering tasks. In this section, we want to evaluate the effectiveness of the proposed FLOATER on these tasks. In particular, we focus on three language understanding benchmark sets, GLUE~\cite{wang2018glue}, RACE~\cite{lai2017race} and SQuAD~\cite{rajpurkar2016squad}.
As mentioned in Section~\ref{sec:compatible}, FLOATER is carefully designed to be compatible with the existing Transformer models. 
Thus, we can utilize pretrained Transformer models to warm-start a FLOATER model easily to be used to finetune on these NLP tasks. 
In this paper, we download the \emph{same} pre-trained RoBERTa model from the official repository as our pretrained Transformer model for all NLP tasks discussed in this section.  
{\bf GLUE Benchmark.}
This benchmark is commonly used to evaluate the language understanding skills of NLP models. Experimental results in Table~\ref{tab:exp-GLUE} show that our FLOATER model outperforms RoBERTa in most datasets, even though the only difference is the choice of positional encoding.
\textbf{RACE benchmark}
Similar to the GLUE benchmark, the RACE benchmark is another widely used test suit for language understanding. Compared with GLUE, each item in RACE contains a significantly longer context, which we believe requires more important to grasp the accurate position information. Like in GLUE benchmark, we finetune the model from the same pretrained RoBERTa checkpoint. We keep the hyperparameters, such as batch size and learning rate, to also be the same. Table~\ref{tab:race-benchmark} shows the experimental results. We again see consistent improvement of FLOATER across all subtasks.

\begin{figure}[htb]
    \centering
    \scalebox{0.67}{\input{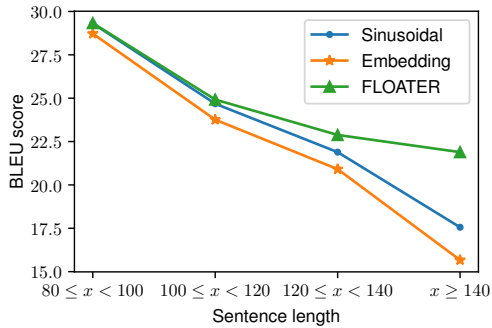}}
    \caption{Comparing BLEU scores of different encoding methods.}
    \label{fig:S2L}
\end{figure}
\begin{table}[htb]
    \centering
    \caption{Performance comparison on WMT14 En-De data and Transformer-base architecture. Both BLUE scores and the number of trainable parameters inside each position encoder are included.}
    \label{tab:rnn-alternative}
    \begin{tabular}{crr}
    \toprule
                       &  BLEU~($\uparrow$) & \#Parameters~($\downarrow$)\\
    \midrule
  FLOATER               &  28.57 &   526.3K \\
  1-layer RNN + scalar &  27.99 &   263.2K \\
  2-layer RNN + scalar &  28.16 &   526.3K \\
  1-layer RNN + vector &  27.99 & 1,050.0K \\
  \bottomrule
    \end{tabular}
\end{table}
\textbf{SQuAD benchmark}
SQuAD benchmark~\cite{rajpurkar2016squad,rajpurkar2018know} is another challenging task to evaluate the question answering skills of NLP models. In this dataset, each item contains a lengthy paragraph containing facts and several questions related to the paragraph. The model needs to predict the range of characters that answer the questions. In SQuAD-v2, the problem becomes more challenging that the questions might be unanswerable by the context. We follow the same data processing script as BERT/RoBERTa for fair comparison; more details about the training process are described in the Supplementary material. The experiment results are presented in Table~\ref{tab:squad}. As we can see, the FLOATER model beats the baseline RoBERTa model consistently across most datasets. The improvement is significant, considering that both models are finetuned from the same pretrained checkpoint.
\subsection{More Discussions and Analysis}
\label{sec:exp-analysis}
\paragraph{How inductive is FLOATER?}
FLOATER is designed to be inductive by a data-driven dynamical model~\eqref{eq:dynamics}. To see how inductive FLOATER is when comparing to existing approaches, we design the following experiment. We first notice that in WMT14 En-De dataset, $98.6\%$ of the training sentences are shorter than $80$ tokens. Based on that, we make a new dataset called \emph{En-De short to long} (or S2L for brevity): this dataset takes all the short sentences ($<80$ tokens) as the training split and all the long sentences ($\ge 80$ tokens) as the testing split. We further divide the testing split to four bins according to the source length fallen in $[80, 100),\ [100, 120),\ [120, 140),\ [140, +\infty)$. BLEU scores are calculated in each bin, and the results are presented in Figure~\ref{fig:S2L}. 

Our FLOATER model performs particularly well on long sentences, even though only short sentences are seen by the model during training. This empirical observation supports our conjecture that FLOATER model is inductive: the dynamics learned from shorter sequences can be appropriately generalized to longer sequences.

\paragraph{Is RNN a good alternative to model the dynamics?} 
Recurrent neural network (RNN) is commonly used to perform sequential modeling. RNN and our continuous dynamical model~\eqref{eq:dynamics} indeed share some commonality. Computing the value at the $i$-th step relies on the results at the $(i-1)$-st step. Further, they all contain trainable parameters, allowing them to adapt to each particular task. Lastly, they can be extrapolated to any length as needed. To see if RNN works equally well, we model the sequence $\{\vp_i\}_{i\in\{1,2,\dots\}}$  with RNN models:
\begin{equation}
    \label{eq:dynamic-RNN}
    \vp_{i+1}=\mathrm{RNN}(\vz_i, \vp_{i}),
\end{equation}
where $\vz_i\in\sR^{d_{\mathrm{in}}}$ is the input to the RNN model at index $i$. Recall in RNN language models, $\vz_i$ is the word embedding or hidden feature of the $i$-th token. In our case, since we apply RNN to learn the encodings as opposed to hidden features, sensible inputs can be scalar value $i$ or vectorized value $\mathrm{Vectorize}(i)$ by sinusoidal encoding. We tried both choices on WMT14 En-De data and found that vectorized value generally works better, though not as good as our FLOATER model. Detailed results can be found in Table~\ref{tab:rnn-alternative}.
\begin{figure*}
    \centering
    \begin{subfigure}{0.45\textwidth}
    \caption{Sinusoidal}\label{fig:sin-cos}
    \includegraphics[width=\linewidth]{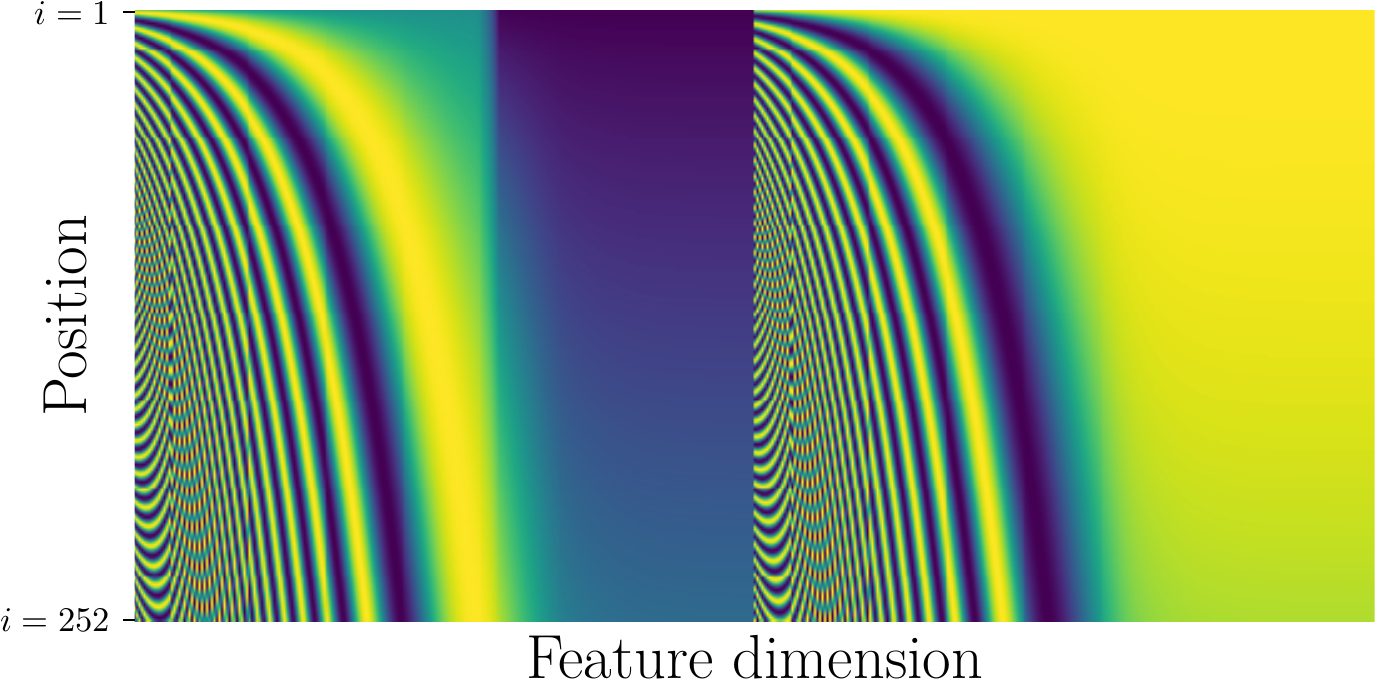}
    \end{subfigure}
    \begin{subfigure}{0.45\textwidth}
    \caption{Position embedding}\label{fig:pos-emb}
    \includegraphics[width=\linewidth]{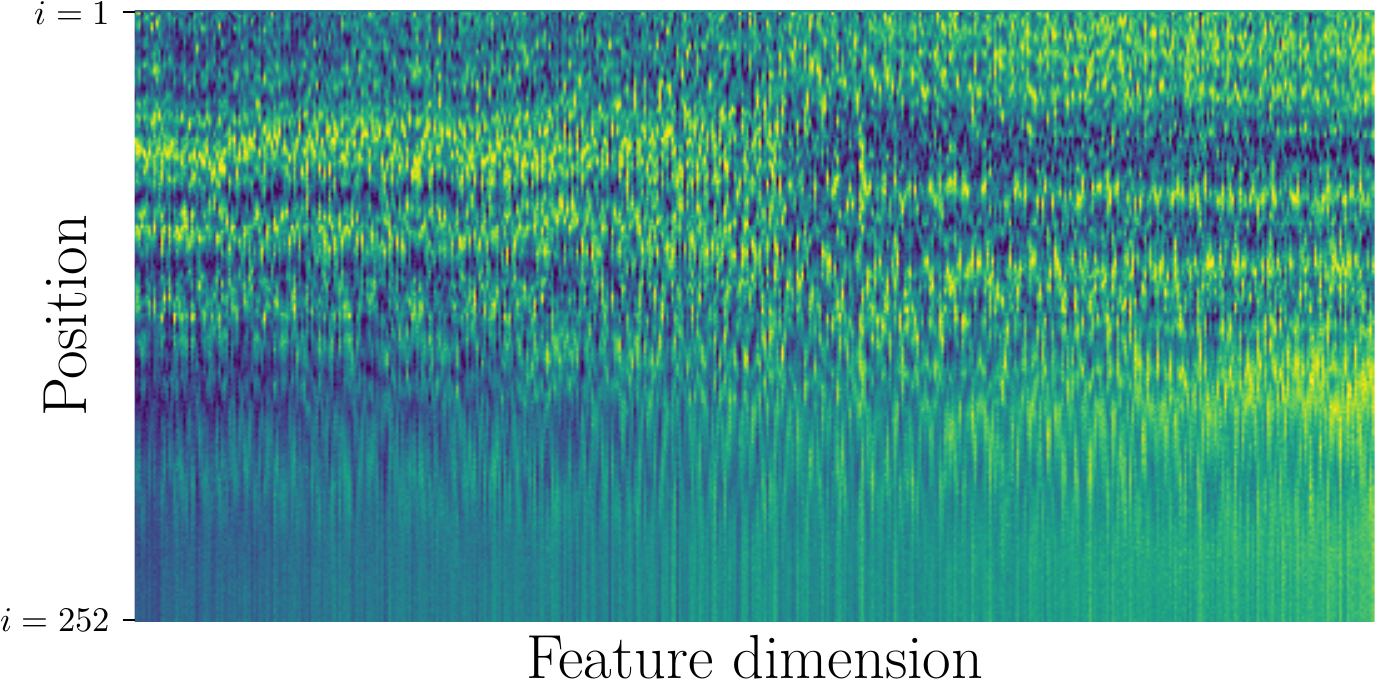}
    \end{subfigure}\\
    \begin{subfigure}{0.45\textwidth}
    \includegraphics[width=\linewidth]{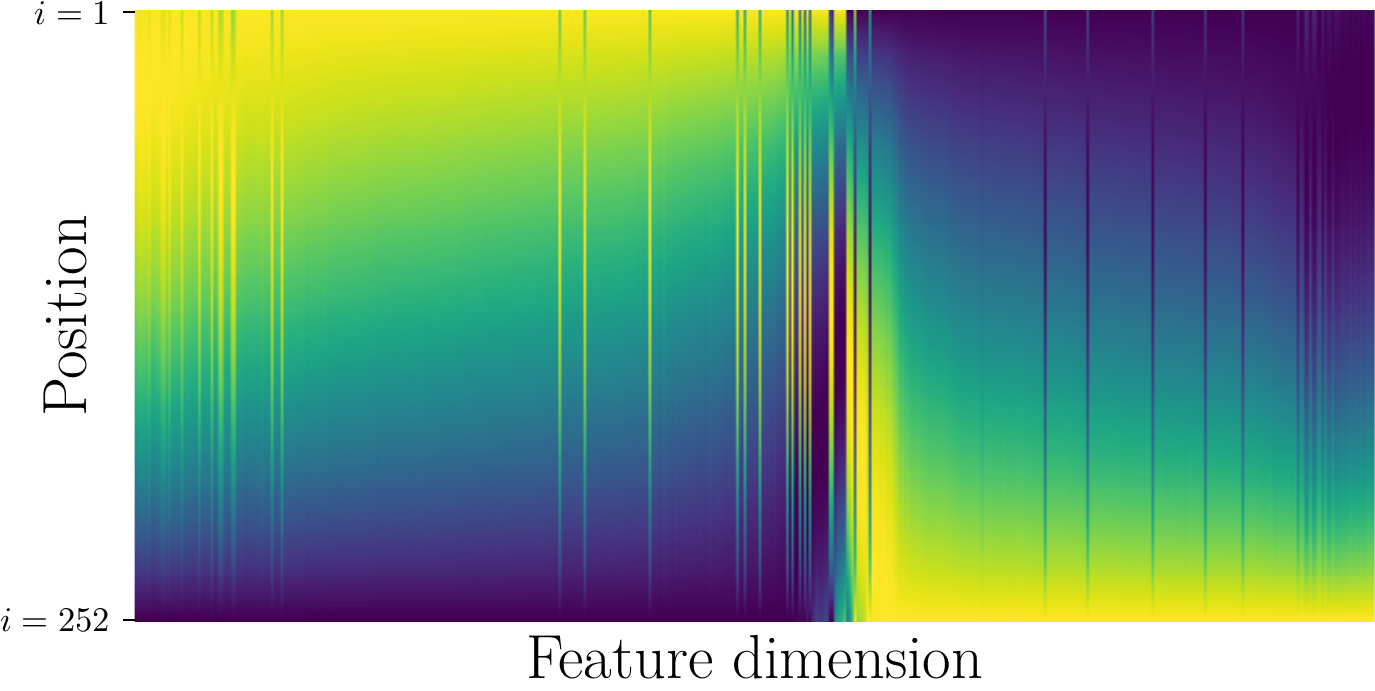}\caption{FLOATER}\label{fig:flower}
    \end{subfigure}
    \begin{subfigure}{0.45\textwidth}
    \includegraphics[width=\linewidth]{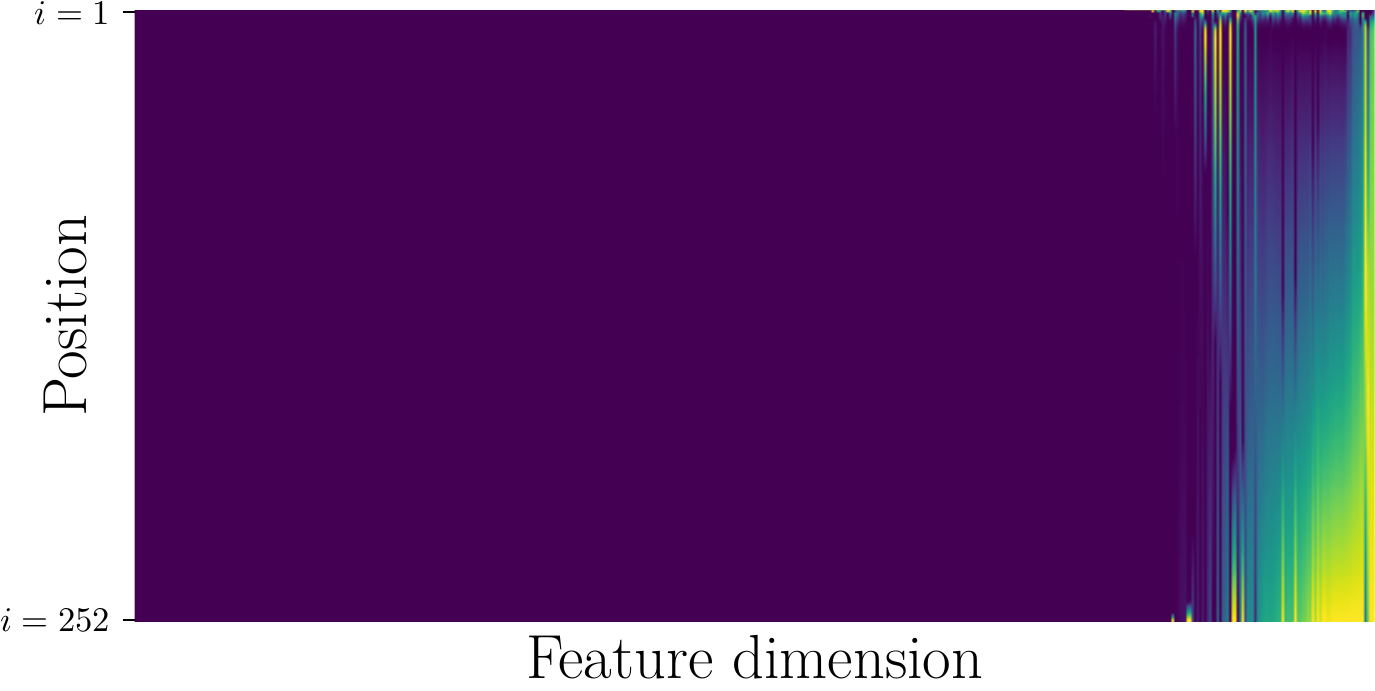}\caption{RNN}\label{fig:rnn}
    \end{subfigure}
    \caption{Visualizing the four different position methods. All models are trained using the Transformer-base architecture and En-De dataset. For better visualization, dimension indices are permuted in Figure~\ref{fig:pos-emb}-\ref{fig:rnn}.}
    \label{fig:visualize-encoding}
\end{figure*}
\paragraph{What does each position encoding look like? }
To better understand how different position encodings affect the sequence modeling, in Figure~\ref{fig:visualize-encoding}, we visualize the position embedding matrix $\vp$ obtained from four different position encoding approaches for the Transformer-base backbone on WMT14 En-De dataset. We can see that sinusoidal encoding (\ref{fig:sin-cos}) is the most structural, while position embedding (\ref{fig:pos-emb}) is quite chaotic. Our FLOATER model learns position representation completely from data, but still exhibits some regularities (\ref{fig:flower}). Finally, the RNN model (\ref{fig:rnn}) fails to extract sufficient positional information, probably due to the vanishing gradient problem. Another finding is that by looking at (\ref{fig:pos-emb}), we observe that the vectors are nearly constant among different large positions (near the bottom of Figure~\ref{fig:pos-emb}, we see patterns of vertical lines with the same color). This phenomenon is due to long sentences in the dataset being scarce, and so the positional information carried by lower indices cannot be extrapolated to higher indices. On the contrary, the dynamical model proposed in this paper enjoys the best of both worlds -- it is adaptive to dataset distribution, and it is inductive to handle sequences with lengths longer than the training split.

\subsection{Remarks on Training and Testing Efficiency}
It is not surprising that during the training time, our flow-based method adds a non-negligible time and memory overhead; this is because solving the Neural ODE precisely involves ${\sim}100$ times forward and backward propagations of the \emph{flow model}. Even though we deliberately designed a small flow model (consisting of only two FFN and one nonlinearity layers), stacking them together still increases training time substantially. To make it possible to train big models, we use the following optimizations:
\begin{itemize}[nosep,leftmargin=1em,labelwidth=*,align=left]
    \item Initialize with pretrained models that do not contain flow-based dynamics, as discussed in Section~\ref{sec:compatible}.
    \item From~\eqref{eq:dynamics}, we know that if $\vh(\cdot)$ is close to zero, then the position information diminishes (derived in appendix). In this way, our model degenerates to the original Transformer. Inspired by this property, we can initialize the FLOATER with smaller weights. Combining with the previous trick, we obtain an informed initialization that incurs lower training loss at the beginning.
    \item We observed that weights in $\vh(\cdot)$ are more stable and easy to train. Thus, we can separate the weights of $\vh(\cdot)$ from the remaining parts of the Transformer model. Concretely, we can 1) cache the positional bias vectors for some iterations without re-computing, 2) update the weights of flow models less frequently than other parts of the Transformer, and 3) update the flow models with a larger learning rate to accelerate convergence.
    \item For the RoBERTa model, we adopt an even more straightforward strategy: we first download a pretrained RoBERTa model, plug in some flow-based encoding layers, and re-train the encoding layers on WikiText-103 dataset for one epoch. When finetuning on GLUE datasets, we can choose to freeze the encoding layers.
\end{itemize}
Combining those tricks, we successfully train our proposed models with only $20$-$30$\% overhead compared to traditional models, and virtually no overhead when finetuning RoBERTa model on GLUE benchmarks. Moreover, there is no overhead during the inference stage if we store the pre-calculated positional bias vectors in the checkpoints.
\section{Conclusions}
In this paper, we have shown that learning position encoding with a dynamical model can be an advantageous approach to improve Transformer models. Our proposed position encoding approach is inductive, data-driven, and parameter efficient. We have also demonstrated the superiority of our proposed model over existing position encoding approaches on various natural language processing tasks such as neural machine translation, language understanding, and question answering tasks.

\bibliography{ref}

\begin{thebibliography}{10}

\bibitem{vaswani2017attention}
Ashish Vaswani, Noam Shazeer, Niki Parmar, Jakob Uszkoreit, Llion Jones,
  Aidan~N Gomez, {\L}ukasz Kaiser, and Illia Polosukhin.
\newblock Attention is all you need.
\newblock In {\em Advances in Neural Information Processing Systems}, pages
  5998--6008, 2017.

\bibitem{devlin2018bert}
Jacob Devlin, Ming-Wei Chang, Kenton Lee, and Kristina Toutanova.
\newblock Bert: Pre-training of deep bidirectional transformers for language
  understanding.
\newblock {\em arXiv preprint arXiv:1810.04805}, 2018.

\bibitem{yang2019xlnet}
Zhilin Yang, Zihang Dai, Yiming Yang, Jaime Carbonell, Ruslan Salakhutdinov,
  and Quoc~V Le.
\newblock Xlnet: Generalized autoregressive pretraining for language
  understanding.
\newblock {\em arXiv preprint arXiv:1906.08237}, 2019.

\bibitem{radford2019language}
Alec Radford, Jeff Wu, Rewon Child, David Luan, Dario Amodei, and Ilya
  Sutskever.
\newblock Language models are unsupervised multitask learners.
\newblock 2019.

\bibitem{lan2019albert}
Zhenzhong Lan, Mingda Chen, Sebastian Goodman, Kevin Gimpel, Piyush Sharma, and
  Radu Soricut.
\newblock Albert: A lite bert for self-supervised learning of language
  representations.
\newblock {\em arXiv preprint arXiv:1909.11942}, 2019.

\bibitem{raffel2019exploring}
Colin Raffel, Noam Shazeer, Adam Roberts, Katherine Lee, Sharan Narang, Michael
  Matena, Yanqi Zhou, Wei Li, and Peter~J Liu.
\newblock Exploring the limits of transfer learning with a unified text-to-text
  transformer.
\newblock {\em arXiv preprint arXiv:1910.10683}, 2019.

\bibitem{yun2019Transformers}
Chulhee Yun, Srinadh Bhojanapalli, Ankit~Singh Rawat, Sashank~J Reddi, and
  Sanjiv Kumar.
\newblock Are transformers universal approximators of sequence-to-sequence
  functions?
\newblock {\em arXiv preprint arXiv:1912.10077}, 2019.

\bibitem{liu2019roberta}
Yinhan Liu, Myle Ott, Naman Goyal, Jingfei Du, Mandar Joshi, Danqi Chen, Omer
  Levy, Mike Lewis, Luke Zettlemoyer, and Veselin Stoyanov.
\newblock Roberta: A robustly optimized bert pretraining approach.
\newblock {\em arXiv preprint arXiv:1907.11692}, 2019.

\bibitem{shaw2018self}
Peter Shaw, Jakob Uszkoreit, and Ashish Vaswani.
\newblock Self-attention with relative position representations.
\newblock In {\em Proceedings of the 2018 Conference of the North American
  Chapter of the Association for Computational Linguistics: Human Language
  Technologies, Volume 2 (Short Papers)}, pages 464--468, 2018.

\bibitem{dehghani2018universal}
Mostafa Dehghani, Stephan Gouws, Oriol Vinyals, Jakob Uszkoreit, and {\L}ukasz
  Kaiser.
\newblock Universal transformers.
\newblock {\em arXiv preprint arXiv:1807.03819}, 2018.

\bibitem{chen2018neural}
Tian~Qi Chen, Yulia Rubanova, Jesse Bettencourt, and David~K Duvenaud.
\newblock Neural ordinary differential equations.
\newblock In {\em Advances in Neural Information Processing Systems}, pages
  6571--6583, 2018.

\bibitem{grathwohl2018ffjord}
Will Grathwohl, Ricky~TQ Chen, Jesse Betterncourt, Ilya Sutskever, and David
  Duvenaud.
\newblock Ffjord: Free-form continuous dynamics for scalable reversible
  generative models.
\newblock {\em arXiv preprint arXiv:1810.01367}, 2018.

\bibitem{clark1964ordinary}
M.~Tenenbaum and H.~Pollard.
\newblock {\em Ordinary Differential Equations: An Elementary Textbook for
  Students of Mathematics, Engineering, and the Sciences}.
\newblock Dover Books on Mathematics. Dover Publications, 1985.

\bibitem{ott2019fairseq}
Myle Ott, Sergey Edunov, Alexei Baevski, Angela Fan, Sam Gross, Nathan Ng,
  David Grangier, and Michael Auli.
\newblock fairseq: A fast, extensible toolkit for sequence modeling.
\newblock {\em arXiv preprint arXiv:1904.01038}, 2019.

\bibitem{lai2017race}
Guokun Lai, Qizhe Xie, Hanxiao Liu, Yiming Yang, and Eduard Hovy.
\newblock Race: Large-scale reading comprehension dataset from examinations.
\newblock {\em arXiv preprint arXiv:1704.04683}, 2017.

\bibitem{wang2018glue}
Alex Wang, Amanpreet Singh, Julian Michael, Felix Hill, Omer Levy, and Samuel~R
  Bowman.
\newblock Glue: A multi-task benchmark and analysis platform for natural
  language understanding.
\newblock {\em arXiv preprint arXiv:1804.07461}, 2018.

\bibitem{rajpurkar2016squad}
Pranav Rajpurkar, Jian Zhang, Konstantin Lopyrev, and Percy Liang.
\newblock Squad: 100,000+ questions for machine comprehension of text.
\newblock {\em arXiv preprint arXiv:1606.05250}, 2016.

\bibitem{rajpurkar2018know}
Pranav Rajpurkar, Robin Jia, and Percy Liang.
\newblock Know what you don't know: Unanswerable questions for squad.
\newblock {\em arXiv preprint arXiv:1806.03822}, 2018.

\bibitem{press1992numerical}
William~H Press, Saul~A Teukolsky, William~T Vetterling, and Brian~P Flannery.
\newblock Numerical recipes in c++.
\newblock {\em The art of scientific computing}, 2:1002, 1992.

\bibitem{ott2018scaling}
Myle Ott, Sergey Edunov, David Grangier, and Michael Auli.
\newblock Scaling neural machine translation.
\newblock {\em arXiv preprint arXiv:1806.00187}, 2018.

\bibitem{merity2016pointer}
Stephen Merity, Caiming Xiong, James Bradbury, and Richard Socher.
\newblock Pointer sentinel mixture models.
\newblock {\em arXiv preprint arXiv:1609.07843}, 2016.

\bibitem{liu2019hierarchical}
Yang Liu and Mirella Lapata.
\newblock Hierarchical transformers for multi-document summarization.
\newblock {\em arXiv preprint arXiv:1905.13164}, 2019.

\bibitem{zhang2019HIBERT}
Xingxing Zhang, Furu Wei, and Ming Zhou.
\newblock {HIBERT:} document level pre-training of hierarchical bidirectional
  transformers for document summarization.
\newblock {\em CoRR}, abs/1905.06566, 2019.

\bibitem{NIPS2018_7892}
Tian~Qi Chen, Yulia Rubanova, Jesse Bettencourt, and David~K Duvenaud.
\newblock Neural ordinary differential equations.
\newblock In S.~Bengio, H.~Wallach, H.~Larochelle, K.~Grauman, N.~Cesa-Bianchi,
  and R.~Garnett, editors, {\em Advances in Neural Information Processing
  Systems 31}, pages 6571--6583. Curran Associates, Inc., 2018.

\end{thebibliography}
\bibliographystyle{unsrt}

\appendix
\section{Training a Neural ODE model in Transformer}
\begin{figure}[htb]
    \centering
    \includegraphics[width=0.7\linewidth]{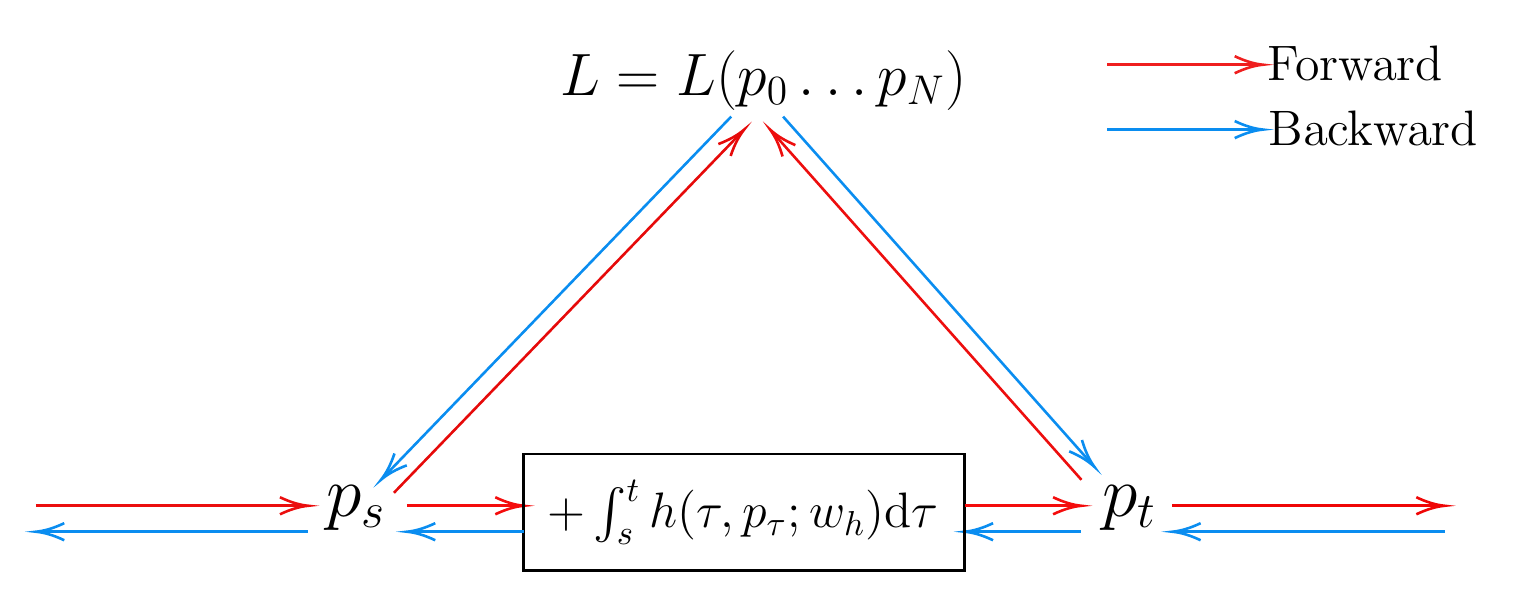}
    \caption{Direction of forward and backward propagation. Here we consider a simplified version where only position encodings $\vp_s$ and $\vp_t$ are in the computational graph.}
    \label{fig:gradient-flow}
\end{figure}

We discuss the details of training the dynamical model $\vh(\tau, \vp_{\tau};\vw_h)$, recall in our FLOWER model, function $\vh$ joins in the computational graph implicitly by generating a sequence of position encoding vectors $\{\vp_1, \vp_2, \dots, \vp_N\}$, conditioning on a freely initialized vector $\vp_0$. The generation steps are computed iteratively as follows (suppose we choose the interval between two consecutive tokens to be $\Delta$) 
\begin{equation}
    \label{eq:p-iterate}
    \begin{aligned}
    \vp_1&=\vp_0+\int_0^{\Delta}\vh(\tau,\vp_{\tau};\vw_h)\dif \tau,\\
    \vp_2&=\vp_1+\int_{\Delta}^{2\Delta}\vh(\tau,\vp_{\tau};\vw_h)\dif \tau,\\
      & \vdotswithin{=} \\
    \vp_N&=\vp_{N-1}+\int_{(N-1)\Delta}^{N\Delta}\vh(\tau,\vp_{\tau};\vw_h)\dif \tau.\\
    \end{aligned}
\end{equation}
Finally, the loss $L$ of this sequence is going to be a function of all position encoding results $L=L(\vp_0,\vp_1,\dots,\vp_N)$, which is further a function of model parameters $\vw_h$. The question is how to calculate the gradient $\frac{\dif L}{\dif \vw_h}$ through backpropagation. This question is fully solved in Neural ODE method~\cite{chen2018neural} with an efficient adjoint ODE solver. To illustrate the principle, we draw a diagram showing the forward and backward propagation in Figure~\ref{fig:gradient-flow}.

From \cite{chen2018neural}, we know that the gradients $\frac{\dif}{\dif \vw_h}L\Big(\vp_s+\int_s^t\vh(\tau,\vp_{\tau};\vw_h)\dif \tau\Big)\triangleq \frac{\dif L}{\dif \vw_h}$ can be computed by
\begin{equation}
    \label{eq:compute-grad}
    \frac{\dif L}{\dif \vw_h}=-\int_t^s\va(\tau)^\intercal \frac{\partial \vh(\tau,\vp_{\tau};\vw_h)}{\partial \vw_h}\dif\tau,
\end{equation}
where $\va(\tau)$ defined in $\tau\in[s,t]$ is called the ``adjoint state'' of ODE, which can be computed by solving another ODE
\begin{equation}
\label{eq:compute-adjoint}
    \frac{\dif \va(\tau)}{\dif \tau}=-\va(\tau)^\intercal \frac{\partial \vh(\tau,\vp_{\tau};\vw_h)}{\partial \vp_{\tau}}.
\end{equation}
Note that the computation of \eqref{eq:compute-adjoint} only involves Jacobian-vector product so it can be efficiently calculated by automatic differentiation.

\section{Implementation details}
\subsection{Settings of ODE solver}
To setup the ODE server, we need to first choose the numerical algorithms~\cite{press1992numerical}. We have different setups for different datasets. For neural machine translation problems (WMT14 En-De and En-Fr), we use the more accurate Runge-Kutta scheme with discretization  step $\frac{\Delta}{5.0}$ to solve the adjoint equation (recall that we set the interval of two neighboring tokens to be $\Delta=0.1$ globally). While for datasets with long sentences such as GLUE and RACE benchmarks, we found that solving the adjoint equation with high order scheme is too slow, in such case we adopt simple midpoint method with discretization  step $\frac{\Delta}{5.0}$, and the gradients are calculated by automatic differentiation rather than adjoint method. The third party implementation of ODE solver can be found at \url{https://github.com/rtqichen/torchdiffeq}.
\subsection{Training NMT tasks}
We run the same preprocessing script provided by fairseq~\cite{ott2019fairseq}, which is also used in ScalingNMT~\cite{ott2018scaling}. With the standard training script, we first successfully reproduce all the results in Transformer paper~\cite{vaswani2017attention}. Based on that we execute the following protocol to get our results:
\begin{enumerate}
    \item Train the original Transformer model for 30 epochs.
    \item Random initialize FLOWER model of same shape configuration. 
    \item Copy tensors from the best performing checkpoint (validation set) to initialize FLOWER model. Initialize weights in the dynamical model with small values.
    \item Half the peak learning rate (e.g. in Transformer-base + En-De, the peak learning rate is changed from $7.0\times 10^{-4}$ to $3.5\times 10^{-4}$).
    \item With the warm-initialized FLOWER checkpoint, retrain on the same dataset for 10 epochs (En-De) or 1 epoch (En-Fr). 
    \item Averaging last 5 checkpoints and compute BLEU score on test split.
\end{enumerate}

\subsection{Training language understanding tasks}
For GLUE/SQuAD/RACE benchmarks, our experiments are all conducted upon RoBERTa, in which both \texttt{base} and \texttt{large} configurations are available. Due to resource constraint (and to show the compatibility to existing models), we initialize our FLOWER model with pretrained RoBERTa, which is similar to NMT task. However, the weights $\vw_h$ in dynamic function $\vh(\tau, \vp_{\tau};\vw_h)$ are not trained in large corpus, given that GLUE/SQuAD/RACE datasets are too small to train dynamics from scratch, we decided to pretrain $\vh$ alone in WikiText103~\cite{merity2016pointer} data using masked language modeling loss. We have found that when we train $\vw_h$ alone, it only takes a few hours (2x Titan V100) and one epoch to convergence. 
\par
Once having the pretrained FLOWER model, we can run following downstream tasks and compare with RoBERTa under the same setting:
\paragraph{GLUE benchmark}consists of eight datasets and each have different hyperparameter settings. For hyperparameters such as learning rate, batch size, training iterations, warm-up iterations, etc., we use the same values recommended by official repository of RoBERTa\footnote{Available at: \url{https://github.com/pytorch/fairseq/blob/master/examples/roberta/README.glue.md}}. 
\paragraph{SQuAD benchmark.} For this benchmark we wrote our own finetuning code because currently there is no official code available. During the implementation process, we mainly refer to the third-party repositories\footnote{ Mainly \url{https://github.com/ecchochan/roberta-squad} and \url{https://github.com/huggingface/transformers}}. We are not able to exactly match the official result reported in RoBERTa paper but quite close (${\sim}0.1$ difference in F1). For our FLOWER model, we use the same hyperparameters as RoBERTa.
\paragraph{RACE benchmark.} This benchmark has the longest context and sequence length. We follow the official training script\footnote{\url{https://github.com/pytorch/fairseq/blob/master/examples/roberta/README.race.md}} and reproduce the result. Similar to other benchmarks, we then repeat the training process using exactly the same training hyperparameters to make a fair comparison. In this benchmark we freeze the weights $\vw_h$ and only finetune the weights of RoBERTa.

\section{Cases suitable for non-equidistant discritization}
Although our model allows continuous values of $s$ and $t$ in \eqref{eq:dynamics}, limiting the scope to text modeling tasks, positions are discrete values as $\{0,1,2,\dots\}$. Once the continuous version of position representation $\vp_t$ is obtained, we simply take the discritized $\{\vp_0,\vp_{\Delta}, \vp_{2\Delta},\dots,\}$ as the actual values to feed into Transformer model, where $\Delta$ is a hyperparameter (\emph{e.g.} $\Delta=0.1$). By choosing positions $t$ equidistantly, we are implicitly assuming the position signal evolves steadily as we go through each token in a sentence. More generally, the dynamics in~\eqref{eq:dynamics} can deal with the case in which positions are not integers $0,1,2,\dots$ etc., but arbitrary monotone increasing series $t_0<t_1<t_2<\dots$ which may not be equidistant. In appendix, we exemplify this general situation with several widely deployed tasks; we regard this as a interesting future direction.
This makes our model particularly suitable for following scenarios yet traditional position representation may not be good at:
\begin{itemize}[nosep,leftmargin=1em,labelwidth=*,align=left]
    \item \textbf{Hierarchical Transformer model}~\cite{liu2019hierarchical,zhang2019HIBERT}. The model is a direct extension of hierarchical RNN and is often used in long document processing. It works by first running a word-level Transformer model on each sentence to extract the sentence embedding, and then applying a sentence-level Transformer scanning through each sentence embedding sequentially. We argue that when processing at the sentence level, it could be better to set the increment of position index $t_{i+1}-t_i$ proportional to the length of the $i$-th sentence. This is because longer sentences tend to carry more information, so $\vp_{i+1}$ is likely to move farther from $\vp_i$.
    \item \textbf{Transformer for time-series events}. As measurement time is continuous, time-series data is another scenario when a continuous position makes more sense than a discrete counterpart. More importantly, to predict the future values by modeling historical values observed at irregular time grids, it is better to consider the length of time horizon between two consecutive measures. A successful previous work is the Latent ODE~\cite{NIPS2018_7892}, except that they use RNN as the backbone, and they model the hidden states rather than position representations with Neural ODE (because RNN itself provides positional bias).
\end{itemize}
In this paper, we are not going to explore the more general cases discussed above. Instead, we decided to leave them as interesting future work.

\end{document}